\documentclass[conference]{IEEEtran}
\IEEEoverridecommandlockouts
\usepackage{cite}
\usepackage{amsmath,amssymb,amsfonts}
\usepackage{graphicx, color}
\usepackage{subcaption}
\usepackage[export]{adjustbox}
\usepackage{textcomp}
\usepackage{url} 
\usepackage[table]{xcolor} 
\usepackage{paralist}
\usepackage{enumitem}
\usepackage{multirow}
\usepackage{hyperref}

\usepackage{algorithmic}
\usepackage[linesnumbered,ruled,vlined]{algorithm2e}
\hypersetup{
    colorlinks=true,
    linkcolor=red, 
    citecolor=blue, 
    urlcolor=magenta    
}
\def\ie{{\em i.e.}}
\def\eg{{\em e.g.}}
\def\BibTeX{{\rm B\kern-.05em{\sc i\kern-.025em b}\kern-.08em
    T\kern-.1667em\lower.7ex\hbox{E}\kern-.125emX}}

\newcommand{\smallurl}[1]{\footnotesize\url{#1}}

\definecolor{baselinecolor}{gray}{.9}
\newcommand{\baseline}[1]{\cellcolor{baselinecolor}{#1}}

\def\sign{\texttt{sign}}

\begin{document}

\title{Robust Light-Weight Facial Affective Behavior Recognition with CLIP}

\author{\IEEEauthorblockN{Li Lin\textsuperscript{1}, Sarah Papabathini\textsuperscript{1}, Xin Wang\textsuperscript{2}, 
Shu Hu\textsuperscript{1}$^*$\thanks{$^*$Corresponding author} }
\IEEEauthorblockA{
{\textsuperscript{1}Purdue University, {\tt \small \{lin1785, Slpapaba, hu968\}@purdue.edu} }\\
\textsuperscript{2}University at Albany, State University of New York {\tt \small xwang56@albany.edu}}
}

\maketitle
\thispagestyle{plain}
\pagestyle{plain}





\begin{abstract}
Human affective behavior analysis aims to delve into human expressions and behaviors to deepen our understanding of human emotions. Basic expression categories (EXPR) and Action Units (AUs) are two essential components in this analysis, which categorize emotions and break down facial movements into elemental units, respectively. Despite advancements, existing approaches in expression classification and AU detection often necessitate complex models and substantial computational resources, limiting their applicability in everyday settings. In this work, we introduce the first lightweight framework 
adept at efficiently tackling both expression classification and AU detection. This framework employs a frozen CLIP image encoder alongside a trainable multilayer perceptron (MLP), enhanced with Conditional Value at Risk (CVaR) for robustness and a loss landscape flattening strategy for improved generalization. Experimental results on the Aff-wild2 dataset demonstrate superior performance in comparison to the baseline while maintaining minimal computational demands, offering a practical solution for affective behavior analysis. The code is available at \url{https://github.com/Purdue-M2/Affective_Behavior_Analysis_M2_PURDUE}.
\end{abstract}

\begin{IEEEkeywords}
Robust, Expression Classification, Action Unit Detection 
\end{IEEEkeywords}

\section{Introduction}
Basic expression categories (EXPR) and Action Units (AUs) are key approaches in affective behavior analysis, a field that is essential for enhancing human-computer interaction by delving into the nuances of human emotions. This exploration is becoming increasingly important for creating more intuitive human-computer interactions. Basic expression categories divide expressions into a limited number of groups according to the emotion categories, \eg, happiness, sadness, etc. AU depicts the local regional movement of faces which can be used as the smallest unit to describe the expression~\cite{prince2015facial}. 

The sixth Competition on Affective Behavior Analysis in-the-wild (ABAW6)~\cite{kollias20246th} is organized with the objective of addressing the challenges associated with human affective behavior analysis. It makes great efforts to construct large-scale multi-modal video datasets Aff-wild and Aff-wild2~\cite{kollias2023abaw2, kollias2023multi,kollias2023abaw,kollias2022abaw,kollias2021analysing,kollias2020analysing,kollias2021distribution, kollias2021affect, kollias2019expression, kollias2019face, kollias2019deep, zafeiriou2017aff}. Introducing these datasets has advanced the field of facial expression analysis in the wild and accelerated the practical implementation of related industries.

Deep learning has emerged as a pivotal technology in affective behavior analysis, demonstrating remarkable success in
enhancing the detection and classification performance. By leveraging complex neural network architectures, deep learning models can automatically learn from vast amounts of face video data, including audio and images, to identify the expression patterns and subtle facial muscle movements. Specifically, in \textit{Expression Classification},  Some works~\cite{zhang2023multi, kim2022facial} have applied extensive fine-tuning to large-scale deep learning models like the Masked Autoencoder (MAE)~\cite{he2022masked} and Swin transformer~\cite{liu2021swin} for expression recognition. Although these methods have delivered promising outcomes, they typically require significant computational resources. Other works~\cite{phan2022expression, zhou2023leveraging} used pre-trained models, but their framework is complex for downstream tasks (\eg, \cite{phan2022expression} trained a Transformer~\cite{vaswani2017attention}).
Some of the above approaches can also be applied to \textit{Action Unit Detection}, and there are methods~\cite{shao2021jaa, Tang_2021_ICCV} specifically designed for AU. Those works use extra annotations related to the face landmarks, which may not be practical for real-world scenarios lacking such detailed annotations.

In this work, we proposed the first lightweight, straightforward framework for expression classification and action unit detection. This framework comprises three modules: frozen Contrastive Language-Image Pre-training (CLIP) \cite{radford2021learning} ViT as feature extractor, trainable 3-layer multilayer perception (MLP), and optimization. We utilize the CLIP ViT L/14~\cite{openclip2021} as an image encoder to capture high-level representations of face images, primarily because it has been trained on an expansive dataset comprising 400M image-text pairs. This extensive training enables the encoder to develop a nuanced understanding of facial features and expressions. The extracted features are processed through the 3-layer MLP, specifically trained for the given task. To bolster the model's precision, particularly in difficult cases, we integrate Conditional Value at Risk (CVaR) into the respective loss functions: cross-entropy for expression classification and binary cross-entropy for action unit detection. The optimization component is designed to smooth the loss landscape, improving the model's generalizability and performance.
Our contributions are summarized as follows:
\begin{enumerate}
    \item We propose the first lightweight efficient framework suitable for expression classification and action unit detection.
    \item We incorporate CVaR into the loss functions, improving the accuracy and reliability of predictions, especially in challenging scenarios for both tasks.
    \item Our method outperforms the baseline in both tasks, as demonstrated in experiments on the Aff-wild2 dataset.
\end{enumerate}


\begin{figure*}[t]
    \centering
    \includegraphics[width=1\textwidth]{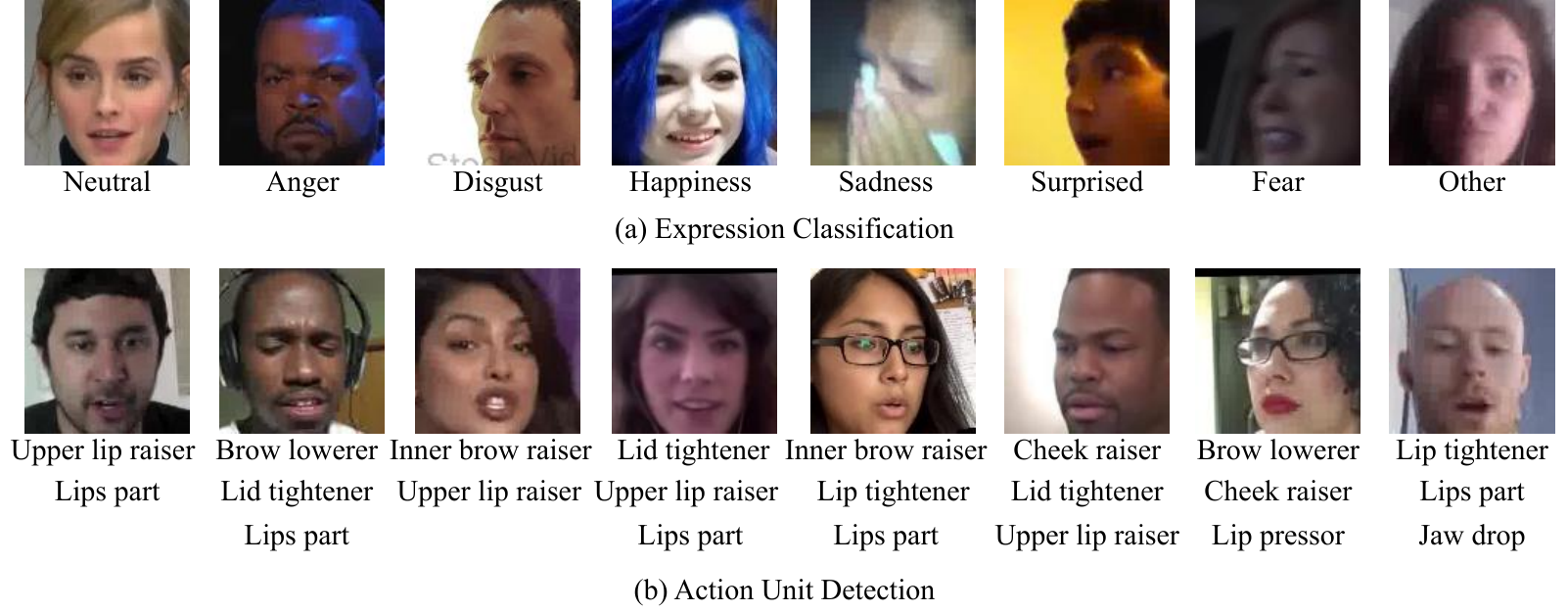}
    \vspace{-5mm}
    \caption{\textit{Displayed samples from Aff-wild2 dataset. \textbf{(a)} Expression Classification is a multi-class task, which has 8 categories in total. \textbf{(b)} Action Unit Detection is a multi-label task, each image contains annotations in terms of 12 AUs.}}
    \vspace{-4mm}
    \label{fig:intro}
\end{figure*}

\section{Method}
Our proposed method can be applied to both \textit{Expression Classification Challenge} and \textit{Action Unit Detection Challenge} \cite{kollias20246th,kollias2023abaw2, kollias2023multi,kollias2023abaw,kollias2022abaw,kollias2021analysing,kollias2020analysing,kollias2021distribution, kollias2021affect, kollias2019expression, kollias2019face, kollias2019deep, zafeiriou2017aff}. The adaptability of our model lies in the configuration of the neurons in the final output layer. Specifically, for the Expression Classification Challenge, the last layer is designed with 8 neurons (8 expressions) with a softmax function. Conversely, for the Action Unit Detection Challenge, this layer is expanded to contain 12 neurons (12 action units) with a sigmoid function. Below, we describe the commonalities that are shared in these two tasks.

\textbf{Feature Space Modeling}. We propose a simple procedure based on features extracted from the image encoder of CLIP ViT L/14~\cite{openclip2021}. CLIP ViT is trained on an extraordinarily large dataset of 400M image-text pairs, so the high-level feature extracted from it is sufficient exposure to the visual world. Additionally, since ViT L/14 has a smaller starting patch size of $14 \times 14$ (compared to other ViT variants), we believe it can also aid in modeling the low-level image details (\ie, texture and edge). Given a dataset $\mathcal{D}= \{(X_i, Y_i)\}_{i=1}^n$ with size $n$, where $X_i$ is the $i$-th sample and $Y_i$ is the $i$-th sample label . Feed CLIP visual encoder with dataset $\mathcal{D}$, and use its final layer to map the training data to their feature representations (of 768 dimensions). We get the resulting feature bank $\mathcal{C} = \{(F_i, Y_i)\}_{i=1}^n$ and further use the feature bank, which is our training set, to design an MLP classifier.

\textbf{MLP Classifier}. After constructing the feature bank, we use those feature embeddings to train our classifier, a straightforward 3-layer Multilayer Perceptron (MLP). To foster a stable learning process and enhance the model's ability to generalize, we incorporate batch normalization after each linear transformation. This is followed by a ReLU activation, allowing the model to effectively capture intricate data patterns. To further combat the risk of overfitting, a dropout layer is included following the activation function. 

\textbf{Objective Function}. To obtain a robust model, we apply a distributionally robust optimization (DRO) technique called \textit{Conditional Value-at-Risk} (CVaR)~\cite{hu2024outlier, ju2024improving, lin2024preserving, hu2023rank, hu2022distributionally, hu2021tkml, hu2022sum, hu2020learning}. By integrating CVaR into the cross-entropy (EC) loss (EC for expression classification, BEC for action unit detection), the model is encouraged to pay more attention to the riskiest predictions. For example, in expression classification, this methodology emphasizes prioritizing expressions such as fear, which not only are less prevalent in the dataset but also closely resemble other expressions, like surprise, making them challenging to distinguish accurately. Addressing these cases is vital because errors here could drastically affect the correct interpretation of emotional states. Similarly, within Action Unit Detection, this enhanced focus on uncertain predictions is pivotal. It ensures the model pays particular attention to subtle or complex facial muscle movements where inaccuracies could lead to significant misinterpretations.

To this end, in what follows, we assume that $\mathcal{C} = \{(F_i, Y_i)\}_{i=1}^n$ consists of i.i.d. samples from a joint distribution $\mathbb{P}$, $F_i$ is the $i$-th data point's feature, and $Y_i$ is the $i$-th point's label. Given some variant of minibatch gradient descent, in these two tasks, we are minimizing the empirical risk of the loss $\mathcal{L}_{avg}(\theta)= \frac{1}{n}\sum_{i=1}^{n}\ell(\theta, F_i, Y_i) \quad \text{for } \theta \in \Theta$,  instead of minimizing the true unknown risk $\mathcal{R}_{avg}(\theta) = \mathbb{E}_{(F,Y) \sim \mathbb{P}}[\ell(\theta; F, Y)]$, where $\ell$ is the individual loss function (\eg, cross-entropy in expression classification, binary cross-entropy in action unit detection) of the model, which has parameters $\theta$ for MLP.

However, the average loss is not robust to the imbalanced data, a common characteristic of the Aff-Wild2 dataset (\eg, Neutral expression has 468,069 images, whereas Fear only contains 19,830 images). In addition, the training data distribution is usually inconsistent with the testing data distribution, which is called the domain shift problem that widely exists in real-world scenarios. For example, training data and testing data come from two different datasets. Therefore, we explore a DRO technique CVaR for handling imbalanced data, which can be formulated as:
\begin{equation*}
    \text{CVaR}_{\alpha}(\theta) = \inf_{\lambda \in \mathbb{R}} \left\{ \lambda + \frac{1}{\alpha} \mathbb{E}_{(F,Y)\sim P} \left[ \left(\ell(\theta; F, Y) - \lambda\right)_+ \right] \right\},
\end{equation*}
As a reminder, $\ell$ is the individual loss function where in expression classification is cross-entropy loss, and in action unit detection is binary cross-entropy loss. $[a]_+ = \max\{0, a\}$ is the hinge function, the conditional value at risk at level $\alpha \in (0,1)$. As $\alpha \to 0$, we are concerned about minimizing the risk of `hard' samples. In contrast, as $\alpha \to 1$, it becomes minimizing the $\mathcal{R}_{avg}(\theta)$.
Inspired by~\cite{ju2024improving}, we can minimize a loss function that aims to minimize an upper bound on the worst-case risk by employing the CVaR. In practice, we minimize an empirical version of $\text{CVaR}_{\alpha}(\theta)$. This gives us the following optimization problem:
\begin{equation}
\begin{aligned}
    \mathcal{L}_{CVaR}(\theta) =  \min_{\lambda \in \mathbb{R}}  \lambda + \frac{1}{\alpha n} \sum_{i=1}^{n} \left[ \ell(\theta; F_i, Y_i) - \lambda \right]_+.
\end{aligned}
\label{eq:learning objective}
\end{equation}

Suppose for a moment that we have obtained the optimal value of $\lambda^*$ in (\ref{eq:learning objective}), then the only training points that contribute to the loss are the `hard' ones with a loss value greater than $\lambda^*$, whereas the `easy' training points with low loss smaller than $\lambda^*$ are ignored. To this end, a robust model for expression classification and action unit detection is obtained. 

\textbf{Optimization}. Last, to further improve the model's generalization capability, we optimize the model by utilizing the sharpness-aware minimization (SAM) method~\cite{foret2020sharpness} to flatten the loss landscape. 
As a reminder, the model's parameters are denoted as $\theta$, flattening is attained by determining an optimal $\epsilon^*$ for perturbing $\theta$ to maximize the loss, formulated as:

\begin{equation}
    \begin{aligned}        
    \epsilon^*&=\arg\max_{\|\epsilon\|_2\leq \gamma}{\mathcal{L}_{CVaR}}\textbf{(}\theta+\epsilon \textbf{)}\\
    &\approx\arg\max_{\|\epsilon\|_2\leq \gamma}\epsilon^\top\nabla_\theta \mathcal{L}_{CVaR}=\gamma\sign(\nabla_\theta \mathcal{L}_{CVaR}),
    \end{aligned}
\label{eq:epsion_star}
\end{equation}
where $\gamma$ is a hyperparameter that controls the perturbation magnitude. The approximation is obtained using first-order Taylor expansion assuming that $\epsilon$ is small. The final equation is obtained by solving a dual norm problem, where $\sign$ represents a sign function and $\nabla_\theta \mathcal{L}_{CVaR}$ being the gradient of $\mathcal{L}_{CVaR}$ with respect to $\theta$. As a result, the model parameters are updated by solving the following problem:
\begin{equation}
    \begin{aligned}
        \min_{\theta} \mathcal{L}_{CVaR}\textbf{(}\theta+\epsilon^*\textbf{)}.
    \end{aligned}
\label{eq:sharpness}
\end{equation}
Perturbation along the gradient norm direction increases the loss value significantly and then makes the model more generalizable while classifying expression and detecting action units. 

\textbf{End-to-end Training}. In practice, we first initialize the model parameters $\theta$ and then randomly select a mini-batch set $C_b$ from $C$, performing the following steps for each iteration on $C_b$:
\begin{compactitem}
    \item Fix $\theta$ and use binary search to find the global optimum of $\lambda$ since (\ref{eq:learning objective}) is convex w.r.t. $\lambda$.
    \item Fix $\lambda$, compute $\epsilon^*$ based on Eq.~(\ref{eq:epsion_star}).
    \item Update $\theta$ based on the gradient approximation for (\ref{eq:sharpness}): $\theta\leftarrow\theta-\beta \nabla_\theta \mathcal{L}_{CVaR}\big|_{\theta+\epsilon^*}$, where $\beta$ is a learning rate.    
\end{compactitem}

\section{Experiments}
\subsection{Experimental Settings}







\textbf{Datasets}. Aff-Wild2~\cite{kollias20246th,kollias2023abaw2, kollias2023multi,kollias2023abaw,kollias2022abaw,kollias2021analysing,kollias2020analysing,kollias2021distribution, kollias2021affect, kollias2019expression, kollias2019face, kollias2019deep, zafeiriou2017aff} is used in both Expression Classification Challenge and the Action Unit Detection Challenge. In the Expression Classification Challenge,  the dataset consists of 548 videos from Aff-Wild2, annotated for the six basic expressions, neutral state, and an `other’ category representing non-basic expressions. Seven videos feature two subjects, both of whom are annotated. In total, annotations are provided for 2,624,160 frames from 437 subjects. The training, validation, and testing sets consist of 248, 70, and 230 videos, respectively. The dataset for the Action Unit Detection Challenge comprises 542 videos annotated for 12 AUs corresponding to the inner and outer brow raiser, the brow lowerer, the cheek raiser, the lid tightened, the upper lip raiser, the lip corner puller and depressor, the lip tightener and pressor, lips part and jaw drop.  The training, validation, and testing sets contain 295, 105, and 142 videos, respectively.

\textbf{Evaluation Metrics}. The performance measure is the average F1 Score. In the Expression Classification Challenge, the average F1 Score across (`macro' F1 score) is 8 categories; in Action Unit Detection, it is across 12 AUs.

\textbf{Baseline Methods}. For our research, we compare our methods with the standard baseline provided by the competition's organizer~\cite{kollias20246th}, referred to as `Official.' In the Expression Classification Challenge, the `Official' baseline uses a VGG16 model that's been pre-trained on the VGGFACE dataset. This model's convolutional layers are set and unchangeable, and it's fine-tuned for our specific task using only its three fully connected layers. It's designed to identify eight facial expressions using a softmax function in the output layer. Data augmentation is applied using a technique called Mixaugment. In the Action Unit Detection Challenge, the baseline is similarly constructed with a fixed VGG16 architecture. Still, it differs in its output layer, which uses a sigmoid function to pinpoint twelve distinct facial action units.

\textbf{Implementation Details}. All experiments are based on the PyTorch and trained with an NVIDIA RTX A6000 GPU. For training, we fix the batch size 32, and epochs 32, and use Adam optimizer with an initial learning rate at $\beta=1e-3$. Additionally, we employ a Cosine Annealing Learning Rate Scheduler to modulate the learning rate adaptively across the training duration, aiming to bolster the model's path to convergence. Experimentally, we find the best $\alpha$ is 0.3, and we set the $\gamma$ to $0.05$.

\begin{table}[t]
\centering
\begin{tabular}{c|c|c}
\hline
Task                                                                                 & Method   & F1 Score      \\ \hline
\multirow{2}{*}{\begin{tabular}[c]{@{}c@{}}Expression\\ Classification\end{tabular}} & Official \cite{kollias20246th} & 0.25          \\
                                                                                     & \baseline{Ours}     & \baseline{\textbf{0.36}} \\ \hline
\multirow{2}{*}{\begin{tabular}[c]{@{}c@{}}Action Unit\\ Detection\end{tabular}}     & Official \cite{kollias20246th} & 0.39          \\
                                                                                     & \baseline{Ours}     & \baseline{\textbf{0.43}} \\ \hline
\end{tabular}
\caption{\textit{Comparison with the baseline method in terms of `macro' F1 score on the validation set in Expression Classification Challenge and Action Unit Detection Challenge, respectively. The best results are shown in \textbf{Bold}.}}
\label{tab:results}
\vspace{-0.1cm}
\end{table}

\begin{figure}[t]
  \centering
  \includegraphics[width=0.8\linewidth]{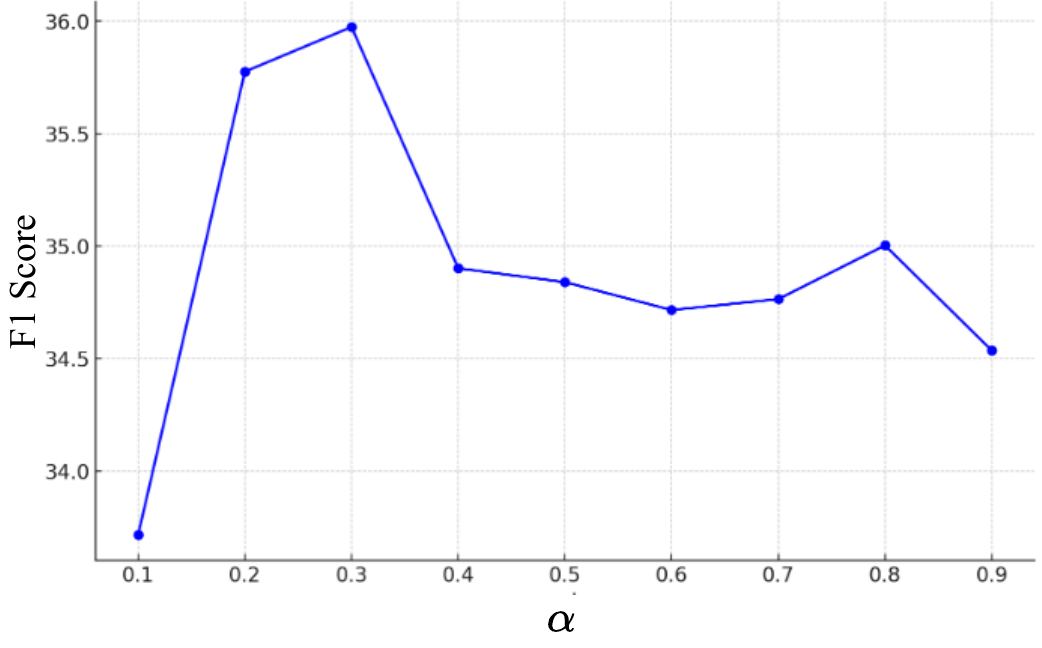}
  \vspace{-2mm}
  \caption{\textit{`Macro' F1 score to different $\alpha$ values in expression classification.}}
  \vspace{-5mm}
  \label{fig:alpha}
\end{figure}

\subsection{Results}
\textbf{Comparison with Baseline}.
Table~\ref{tab:results} presents a comparative evaluation of our method against the baseline technique regarding the ‘macro’ F1 score for both the Expression Classification Challenge and the Action Unit Detection Challenge. Our method outperforms the official baseline, achieving a ‘macro’ F1 score improvement of 11\% in the Expression Classification Challenge, rising from 0.25 to 0.36. In the Action Unit Detection Challenge, our approach exhibits a superior performance as well, enhancing the ‘macro’ F1 score by 4\%, from 0.39 to 0.43. These results underscore the effectiveness of our proposed method, particularly in its ability to more accurately classify expressions and detect action units, highlighting our method's superior performance in both challenges.

\textbf{Sensitivity Analysis}. Fig.\ref{fig:alpha} illustrates the `macro' F1 score across different $\alpha$ values in Eq. (\ref{eq:learning objective}) of expression classification. The $\alpha$ in CVaR controls how much the model prioritizes the tail end of the loss distribution, which corresponds to the most challenging cases during training. The optimal performance observed at $\alpha = 0.3$ implies that targeting the most difficult 30\% of the data enhances the model's ability to handle critical instances, yielding the highest F1 Score. Beyond this point, as $\alpha$ increases, there is a noticeable decline in performance, signaling that too much focus on the tail may lead to over-prioritization of complex examples, potentially reducing overall model effectiveness.

\section{Conclusion}
Existing methods for human affective behavior analysis utilizing complex deep learning models are often associated with high computational costs, limiting their feasibility and scalability in real-world applications. To overcome this limitation, we introduce the first efficient lightweight that employs a frozen CLIP image encoder and a trainable MLP, augmented with CVaR and a loss landscape flattening strategy. The CVaR integration bolsters our model's robustness, while the loss flattening strategy enhances generalization. Experimental results showcase that our framework not only outperforms the baseline but does so with a streamlined and resource-efficient architecture, establishing a new benchmark for efficient yet effective affective behavior analysis.

\smallskip
\smallskip
\noindent\textbf{Acknowledgments.} This work is supported by the U.S. National Science Foundation (NSF) under grant IIS-2434967 and the National Artificial Intelligence Research Resource (NAIRR) Pilot and TACC Lonestar6. Sarah Papabathini is supported by the Office of Naval Research (ONR) under award N00014-21-1-2691.
The views, opinions and/or findings expressed are those of the author and should not be interpreted as representing the official views or policies of NSF, NAIRR Pilot, and ONR.




\bibliographystyle{ieeetr}
\bibliography{main}

\end{document}